\documentclass{article}

\usepackage{arxiv}

\usepackage[utf8]{inputenc} 
\usepackage{hyperref}       
\usepackage{url}            
\usepackage{booktabs}       
\usepackage{amsfonts}       
\usepackage{nicefrac}       
\usepackage{microtype}      
\usepackage{lipsum}		
\usepackage{graphicx}
\usepackage{natbib}
\usepackage{doi}
\usepackage{algorithm}
\usepackage{algpseudocode}
\usepackage{amsmath}
\usepackage{array}
\usepackage[caption=false,font=normalsize,labelfont=sf,textfont=sf]{subfig}
\usepackage{textcomp}
\usepackage{stfloats}
\usepackage{url}
\usepackage{verbatim}
\usepackage{graphicx}
\usepackage{tabularx}
\usepackage{booktabs}

\title{1 BIT IS ALL WE NEED: Binary Normalized Neural Networks}

\author{ \href{https://orcid.org/0000-0001-6632-2692}{\includegraphics[scale=0.06]{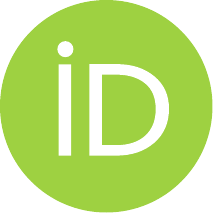}\hspace{1mm}Eduardo L. L. Cabral} \\
	 Mauá Institute of Technology\\
     São Caetano do Sul, São Paulo, SP, Brazil
     \\ Nuclear and Energy Research Institute
     \\São Paulo, SP, Brazil\\
	\texttt{elcabral@maua.br} \\
	\And
	\href{https://orcid.org/0000-0002-4714-287X}{\includegraphics[scale=0.06]{orcid.pdf}\hspace{1mm}Paulo Pirozelli} \\
	 Mauá Institute of Technology\\
     São Caetano do Sul, São Paulo, SP, Brazil \\
	\texttt{paulopirozelli@gmail.com} \\
    	\And
	\href{https://orcid.org/0000-0002-3947-0590}{\includegraphics[scale=0.06]{orcid.pdf}\hspace{1mm}Larissa Driemeier} \\
	 Department of Mechatronics and Mechanical Systems Engineering \\
     Polytechnic School – University of São Paulo\\
     São Paulo, SP, Brazil\\
	\texttt{driemeie@usp.br }} 
\renewcommand{\shorttitle}{1 BIT IS ALL WE NEED}

\hypersetup{
pdftitle={1 BIT IS ALL WE NEED: Binary Normalized Neural Networks},
pdfsubject={},
pdfauthor={Eduardo L.L.~Cabral, Paulo Pirozelli, Larissa Driemeier},
pdfkeywords={Neural networks, Binary parameters, Binary normalized layers},
}

\begin{document}
\maketitle

\begin{abstract}
The increasing size of large neural network models, specifically language models and foundational image models, poses deployment challenges, prompting efforts to reduce memory requirements and enhance computational efficiency. These efforts are critical to ensure practical deployment and effective utilization of these models across various applications. In this work, a novel type of neural network layers and models is developed that uses only single-bit parameters. In this novel type of models all parameters of all layers, including kernel weights and biases, only have values equal to zero or one. This novel type of models uses layers named as binary normalized layer. These binary normalized layers can be of any type, such as fully connected, convolutional, attention, etc., and they consist of slight variations of the corresponding conventional layers. To show the effectiveness of the binary normalized layers, two different models are configured to solve a multiclass image classification problem and a language decoder to predict the next token of a sequence. The model to solve the image classification has convolutional and fully connected layers, and the language model is composed of transformer blocks with multi-head attention. The results show that models with binary normalized layers present almost the same results obtained by equivalent models with real 32-bit parameters. The binary normalized layers allow to develop models that use 32 times less memory than current models and have equivalent performance. Besides, the binary normalized layers can be easily implemented on current computers using 1-bit arrays, and do not require the development of dedicated electronic hardware. This novel type of layers opens a new era for large neural network models with reduced memory requirements that can be deployed using simple and cheap hardware, such as mobile devices or only cpus.
\end{abstract}


\section{Introduction}
Recent advances in machine learning techniques and hardware have enabled extraordinary performance in a wide range of applications — from traditional tasks such as pattern recognition and natural language processing to more complex domains, including autonomous control systems and the discovery of new materials. These developments have been made possible thanks to improved neural network architectures, increased computational power, and the availability of large datasets \cite{goodfellow2016deep}.

However, despite such impressive progress, current artificial intelligence models still face serious limitations when applied to embedded systems. Most state-of-the-art AI solutions rely heavily on cloud computing infrastructure and high-performance specialized hardware. Deep neural networks, such as those used for image classification, often require billions of floating point operations to process a single input sample \cite{Henzinger2021Scalable}. The increasing size of large-scale AI models, especially language models and foundational image models, introduces significant deployment challenges. These models require substantial computational resources, energy, and memory requirements that are difficult to satisfy outside data centers.

This dependency makes it impractical to implement these models in systems with limited computational resources, especially in contexts where connectivity is restricted or nonexistent. Applications in isolated environments, such as underwater, underground, aerospace, or agricultural systems, face severe communication challenges, including physical interference, high latency, and limited bandwidth. As discussed in \cite{Plastiras2018}, local data processing becomes an evolutionary necessity to ensure the autonomy and responsiveness of such systems, particularly where rapid decision making is required under resource constraints.


To address these constraints, quantization has emerged as a key technique for optimizing neural networks in resource-limited environments \cite{Henzinger2021Scalable}. Instead of relying on high-precision floating-point arithmetic (e.g., 32-bit), quantization uses lower-bit integer formats. It typically operates in the 2 to 8 bit range, which is widely adopted in industry with minimal impact on accuracy \cite{Henzinger2021Scalable}. This approach enables significantly reduced memory usage and bandwidth requirements, with compression ratios ranging from $35\times$ to $49\times$, as shown by \cite{han2015deepcompression}. It also provides substantial speed-ups, achieving up to $3$ times faster execution on standard CPUs and up to $10$ times on specialized fixed point hardware such as Qualcomm DSPs with HVX support \cite{jacob2018quantization}. Furthermore, energy efficiency is greatly improved, which is especially critical for mobile and edge devices \cite{krishnamoorthi2018quantization, banner2019post}.

However, quantization presents challenges, particularly in preserving accuracy due to the non-differentiable nature of quantization functions and the sensitivity to ranges of varying values in weights and activations \cite{hubara2017quantized}. To address these issues, several techniques have been developed. 

Post-Training Quantization (PTQ) and Quantization-Aware Training (QAT) are two main strategies to reduce the precision of parameters of neural network models. PTQ applies quantization after a model has been fully trained in high precision (typically 32-bit floating point), aiming to reduce model size and inference latency without retraining. In contrast, QAT simulates quantization effects during training, allowing the model to learn and adapt to the information loss introduced by low-precision representations. As a result, QAT generally achieves significantly higher accuracy than PTQ, particularly at lower bit-widths.

Several key studies have shaped the development of both PTQ and QAT. \cite{jacob2018quantization} presented a foundational quantization scheme that has been widely adopted in TensorFlow Lite, offering practical guidelines for deploying low-precision models in real-world applications. \cite{krishnamoorthi2018quantization} provided a comprehensive white paper that systematically covers the principles and implementation details of both PTQ and QAT, serving as a central reference for researchers. For low-bit PTQ, \cite{banner2019post} proposed ACIQ (Analytical Clipping for Integer Quantization), a method for minimizing quantization error without retraining, particularly effective in the 4-bit setting.

Regarding QAT specifically, \cite{hubara2017quantized} introduced a seminal approach to training Quantized Neural Networks (QNNs) with low-precision weights and activations using the Straight-Through Estimator (STE) to handle non-differentiability. As proposed in \cite{jacob2018quantization}, weights and biases are stored in 32-bit floating-point format to allow precise updates, but are quantized during the forward pass to simulate low-precision inference. Backpropagation still occurs using full-precision gradients, and the STE is employed to handle the non-differentiability of quantization. This dual approach ensures that small gradient updates are not lost, which would happen if parameters were permanently quantized during training. Once training is complete, only the low-precision parameters are retained for efficient inference. In \cite{oord2018} the authors discussed the Vector Quantised-Variational AutoEncoder (VQ-VAE), a generative model that learns discrete latent representations. In \cite{choi2019accurate} the authors proposed PACT (Parameterized Clipping Activation), a method that improves activation quantization by learning optimal clipping thresholds during training. The field has also advanced with techniques such as SAWB (Statistical Aware Weight Binning) for weight quantization, which selects quantization ranges based on the weight distribution to reduce accuracy loss in low-precision networks. \cite{Zhuang2018TowardsEffective} proposed training strategies for quantized networks, including two-stage optimization, progressive quantization, and the use of a full-precision teacher model to guide learning and improve final performance. 

Recently, \cite{Cabral2025} explored the trade-offs involved in using low-bit representations for network weights. Their study demonstrates that 2.32-bit weights—corresponding to five discrete levels—can provide a favorable balance between memory efficiency and model accuracy. They also observe that low-resolution models with fewer parameters may need more training epochs to reach the accuracy of 32-bit models, while larger models can achieve similar performance within typical training configurations.

Recent progress includes BitNet, a transformer architecture tailored for large language models using binary weights and low-precision activations while preserving full-precision states for optimizers and gradients. Its variant, BitNet 1.58, introduces ternary weights (–1, 0, +1), offering comparable performance to 16-bit models with reduced memory footprint, lower latency, and improved energy efficiency.
 
Particularly, 1-bit quantization refers to the technique of reducing the numerical precision of neural network weights and/or activations to just one bit, typically representing values as +1 or –1 \cite{hubara2017quantized}. This extreme form of quantization, known as Binary Neural Networks (BNNs), enables the replacement of costly floating-point operations with highly efficient bitwise operations such as XNOR and bit-counting. The primary goal is to drastically reduce memory usage, energy consumption, and computational complexity, making deep learning models suitable for deployment on low-power, resource-constrained devices. In theory, 1-bit quantization can achieve up to 32× compression over 32-bit floating-point parameters and similarly large energy savings.

Early works by \cite{Hwang2014FixedpointFN,Courbariaux2015BinaryConnectTD} demonstrated the feasibility of training deep models with binary weights. \cite{Hubara2016BinarizedNN} extended this to both weights and activations, training BNNs on datasets like MNIST, CIFAR-10, SVHN, and even ImageNet. Later, \cite{Rastegari2016XNORNet} introduced XNOR-Net, which used a scaling factor to improve the performance of binarized layers, achieving competitive top-1 accuracy on ImageNet. Despite these advances, 1-bit models still suffer from accuracy degradation on complex tasks. For instance,  BNN proposed by \cite{Hubara2016BinarizedNN} achieved 41.8\% top-1 accuracy on AlexNet with ImageNet, while XNOR-Net reached 44.2\%.

In this work, we propose a novel class of neural network models built entirely with single-bit parameters, using binary normalized layers. Unlike traditional models that rely on 32-bit floating-point precision, our approach constrains all layer parameters—including kernel weights and biases—to a single bit of resolution. The binary normalized layer concept is versatile and can be applied across various architectures such as fully connected, convolutional, and attention layers. To demonstrate their effectiveness, we apply these layers to two distinct problems: multiclass image classification using a convolutional binary model, and a language decoder for next-token prediction in language sequences using a binary transformer model. Our results show that these binary models achieve performance comparable to their full-precision 32-bit counterparts, without exhibiting common training instabilities associated with low-resolution parameter networks. This significant memory reduction, of up to 32 times less than conventional models, combined with their straightforward implementation on standard hardware using 1-bit arrays, opens the door to deploying large-scale neural networks on resource-limited platforms such as mobile devices and CPUs. Moreover, the reduced memory footprint allows scaling to larger models, making advanced AI feasible on embedded systems.

Section \ref{sec:BinaryNormalizedLayers} of this paper outlines the binary normalized layers, Sections \ref{sec:ConvolutionalModel} details the convolutional model and the image dataset used to train this model for a multiclass classification problem, Section \ref{sec:LanguageModel} shows the language models and the dataset used to train this model to predict the next token. The results of the both models are compared with results obtained with the corresponding conventional models with 32-bit parameters. Finally, Section \ref{sec:Conclusions} summarizes the conclusions.

\section{Binary normalized layers}
\label{sec:BinaryNormalizedLayers}

\noindent In our binary normalized layers, including the kernel and bias, each parameter exists in two forms simultaneously during training: a full-precision 32-bit floating-point value ($p$) used for gradient updates, and its binarized counterpart ($p_b$) used for forward computations. The quantization process is straightforward and is performed as follows,
\begin{equation}
p_b = \begin{cases} 
1, & \text{if } p > p_{mean} \\
0, & \text{if } p \leq p_{mean} 
\end{cases}
\label{eq:quantization}
\end{equation}

\noindent where $p_{mean}$ is the mean value of the parameters of the layer.

The 32-bit parameters ($p$) are essential because gradient updates during backpropagation are typically very small ($10^{-4}$ to $10^{-2}$), and would be completely lost if parameters were permanently binarized during training. During the forward pass, we use the binary parameters ($p_b$) to compute activations, but during backpropagation the full-precision parameters ($p$) are used to calculate the gradients.

This dual representation approach, inspired by VQ-VAE~\cite{oord2018} and related to QAT principles~\cite{jacob2018quantization}, allows effective training while ultimately delivering the benefits of 1-bit inference. After training completes, we discard the 32-bit parameters and retain only the 1-bit $p_b$ parameters for deployment.

Obviously, training the model with the help of 32-bit parameters requires a large amount of memory, the same as that required in conventional models. But the final trained model only has 1-bit parameters, thus demanding a much smaller amount of memory. Note that to avoid using 32-bit parameters during training, current neural network training methods based on gradient descent could not be used, and a new training method for neural networks with 1-bit parameters would have to be developed.

The fundamental operation in any neural network layer consists of multiplying input data by the layer's weights (kernel), adding biases, and applying an activation function. This linear transformation followed by nonlinear activation enables the network to learn complex patterns. However, when weights are constrained to only zeros and ones, this transformation exhibits two critical limitations. First, it disproportionately amplifies large positive and negative input values while suppressing small ones, making it inadequate for extracting complex features from the input data. Second, the binary nature significantly intensifies both vanishing and exploding gradient problems during backpropagation.

One effective way to address the challenges posed by using 1-bit parameters is to normalize the output of the linear transformation before applying the activation function. While this strategy is commonly employed in modern architectures through normalization layers, it becomes particularly crucial in the context of binary-weighted networks. In these models, normalization not only stabilizes training but also compensates for the severe limitations introduced by extreme quantization, enabling effective learning despite the low resolution of the parameters.

The motivation for this normalization shares similarities with its role in conventional networks but takes on greater importance in binary models due to their restricted representational capacity and sensitivity to input scale. Specifically:
\begin{itemize}
\item \textbf{Equalizing feature influence:} when input features vary in scale, the limited expressiveness of low-precision weights prevents the model from compensating for dominant features. Normalization ensures that all inputs contribute more equally to learning.

\item \textbf{Improving convergence stability:} scale discrepancies in the input can lead to unstable or inefficient optimization. Normalization mitigates this by aligning feature scales, facilitating smoother convergence.

\item \textbf{Controlling gradient magnitudes:} quantized parameters make gradient updates more sensitive to input scale. Normalizing inputs helps keep gradients within a stable range, avoiding saturation or stagnation during training.

\item \textbf{Avoiding biased learning:} when features have unequal numerical ranges, the model may overemphasize those with larger absolute values. Normalization enforces fairer treatment of all features, improving generalization.

\item \textbf{Mitigating vanishing/exploding gradients:} limited-precision models are more prone to unstable gradient propagation. Normalization helps maintain consistent signal flow across layers, especially in deeper networks.
\end{itemize}

Four types of normalized binary layers are implemented and used in different models: fully connected, convolutional, attention and embedding layers. These layers are described in the sections that follow.

\subsection{Binary normalized fully connected layer (BNFCL)}
Algorithm \ref{alg:01} illustrates the forward propagation process in a binary normalized fully connected layer (BNFCL). In Algorithm \ref{alg:01}, \texttt{Quant} denotes the function that performs weight quantization (defined by equation 1); \texttt{NoGradient} is a placeholder function that prevents gradient calculation for its argument during model training; \texttt{Normalize} is the normalization function; \texttt{Activation} represents the chosen activation function for the layer; and trainable is a flag to indicate if the layer is in training or predicting. Note that the Normalize function normalizes the features of each example so that it has zero mean and unit standard deviation.

\begin{algorithm*}
\caption{Forward propagation calculation process in a binary normalized fully connected layer (BNFCL)}
\begin{algorithmic}[1]
\Require Input $x$, weights $W$, bias $b$, flag $trainable$, activation function
\Ensure Activations $a$
\If{$trainable$}
    \State Quantize kernel for training: $W_q = W + \texttt{NoGradient}(\texttt{Quant}(W) - W)$
    \State Quantize bias for training: $b_q = b + \texttt{NoGradient}(\texttt{Quant}(b) - b)$
\Else
    \State Quantize kernel for inference: $W_q = \texttt{Quant}(W)$
    \State Quantize bias for inference: $b_q = \texttt{Quant}(b)$
\EndIf
\State Apply linear transformation: $z = W_q x + b_q$
\State Normalize features of each example: $z = \texttt{Normalize}(z)$
\State Calculate activations: $a = \texttt{Activation}(z)$
\State \Return $a$
\end{algorithmic}
\label{alg:01}
\end{algorithm*}

In Algorithm \ref{alg:01}, $W$ and $W_q$ represent respectively the 32-bit float and binary kernel weights, while $b$ and $b_q$ are the corresponding 32-bit float and binary bias vectors. The arrays $W$ and $W_q$ have shape ($n_x$, $n_{units}$), where $n_x$ is the number of input features and $n_{units}$ is the number of neurons in the layer. The bias vectors $b$ and $b_q$ are one-dimensional, each containing $n_{units}$ elements.

During both training and inference, the binary weights are used to compute the layer activations. However, during training, when $trainable==True$, the gradients are computed and applied to the 32-bit floating-point weight matrix $W$ and bias vector $b$. These high-precision values are retained and updated throughout training, ensuring that no information is lost during the optimization process. This approach enables parameter updates using full-precision values while still performing forward passes with quantized weights. This scheme is adapted from the method proposed in \cite{alcorn2023}, and is similar in spirit to Quantization Aware Training (QAT) \cite{jacob2018quantization}.

It is important to observe that after training only the quantized 1-bit parameters ($W_q$ and $b_q$) are need for inference and the calculations performed in the BNFC layer are modified according to Algorithm \ref{alg:02}.

\begin{algorithm*}
\caption{Forward propagation calculation process in a binary normalized fully connected layer (BNFCL) after training}
\begin{algorithmic}[1]
\Require Input $x$, binary eights $W_q$, binary bias $b_q$, activation function
\Ensure Activations $a$
\State Apply linear transformation: $z = W_q x + b_q$
\State Normalize features of each example: $z = \texttt{Normalize}(z)$
\State Calculate activations: $a = \texttt{Activation}(z)$
\State \Return $a$
\end{algorithmic}
\label{alg:02}
\end{algorithm*}

\subsection{Binary normalized convolutional layer (BNCVL)}
The only difference between the binary normalized convolutional layer (BNCVL) and the binary normalized fully connected layer (BNFCL) is that a convolution operation is used between the filters (kernel with binary weights) and the input tensor of the layer, rather than a simple matrix multiplication. Equation (2) performs a convolution operation in calculating activations in a BNCV layer.

\begin{equation}
z = \texttt{Conv}(W_q x) + b_q
\label{eq:z_def}
\end{equation}
where $\texttt{Conv}(W_q x)$ performs the convolution of $x$ by $W_q$. In this case $W_q$ is the binary kernel parameters of the layer, which is a four-dimensional array with dimensions ($n_H, n_W, n_C, n_F$), where $n_H$, $n_W$ and $n_C$ are respectively the height, the width and the number of channels of the input data, and $n_F$ is the number of filters used in the convolution layer, and $b_q$ is the binary bias vector with $n_F$ elements. The forward propagation process in a binary normalized convolutional layer (BNCVL) is defined in Algorithm \ref{alg:03}. It should be noted that in the BNCVL, equation \ref{eq:z_def} replaces the linear transformation in Algorithm \ref{alg:01}.

\begin{algorithm*}
\caption{Forward propagation calculation process in a binary normalized convolutional layer (BNCVL)}
\begin{algorithmic}[1]
\Require Input $x$, weights $W$, bias $b$, flag $trainable$, activation function
\Ensure Activations a
\If{$trainable$}
    \State Quantize kernel for training: $W_q = W + \texttt{NoGradient}(\texttt{Quant}(W) - W)$
    \State Quantize bias for training: $b_q = b + \texttt{NoGradient}(\texttt{Quant}(b) - b)$
\Else
    \State Quantize kernel for inference: $W_q = \texttt{Quant}(W)$
    \State Quantize bias for inference: $b_q = \texttt{Quant}(b)$
\EndIf{alg:03}
\State Calculate convolution and add the bias: $z = \texttt{Conv}(W_q x) + b_q$
\State Normalize features of each example: $z = \texttt{Normalize}(z)$
\State Calculate activations: $a = \texttt{Activation}(z)$
\State \Return $a$
\end{algorithmic}
\label{alg:03}
\end{algorithm*}

It is import to observe that after training only the quantized 1-bit parameters are need for inference and the calculations performed in the BNCV layer are modified similarly to what it is done for the BNFC layer in Algorithm \ref{alg:02}.

\subsection{Binary embedding layer (BEMBL)}
Algorithm \ref{alg:o4} presents the calculations performed in a binary token and position embedding layer (BEMBL). In Algorithm \ref{alg:o4}, $seq$ is the input of the layer which consists of a sequence of tokens with maximum length equal to $max\_len$, $emb\_dim$ is the dimension of the embedding vectors for each token, and $vocab\_size$ is the number of tokens in the dictionary.

\begin{algorithm*}
\caption{Forward propagation calculation process in a binary normalized embedding layer (BEMB)}
\begin{algorithmic}[1]
\Require Input $seq$, sequence maximum length $max\_len$, embedding dimension $emb\_dim$, vocabulary size $vocab\_size$
\Ensure Token and position embeddings $tk\_pos\_emb$
\State Create linear position vector varying from 0 to $max\_len$: 
\State \ \ \ \ $pos =\texttt{LinearSeq(min\_value}=0,\texttt{max\_value}$ $= max\_len$)
\State One-hot codification of position vector: $one\_hot\_pos = \texttt{ToCategorical}(pos, max\_len)$
\State Embedding codification of positions: $pos\_emb = \texttt{BNFCL(units}=emb\_dim)(one\_hot\_pos)$
\State One-hot codification of token sequence: $one\_hot\_tk = \texttt{ToCategorical}(seq, vocab\_size)$
\State Embedding codification of token sequence: 
\State \ \ \ \ $tk\_emb = \texttt{BNFCL(units}=emb\_dim)(one\_hot\_tk)$
\State Add token and position embeddings: $tk\_pos\_emb = tk\_emb + pos\_emb$
\State \Return $tk\_pos\_emb$
\end{algorithmic}
\label{alg:o4}
\end{algorithm*}

In Algorithm \ref{alg:o4} \texttt{LinearSeq(min\_value, max\_value)} denotes a function that creates a vector varying linearly from the minimum value (\texttt{min\_value}) to the maximum value minus one (\texttt{max\_value-1}); \texttt{BNFCL(units)} is a binary normalized fully connected layer with number of units equal to units, as described in Algorithm \ref{alg:01}; and \texttt{ToCategorical(s, n\_cats)} is a function that performs one-hot codification of a sequence \texttt{s} with \texttt{n\_cats} categories. Note that the activation functions of both BNFCL layers are linear.

\subsection{Binary transformer block with binary attention layer (BTFB)}
A transformer block with its attention mechanism uses only embedding layers, fully connected layers, and normalization layers. The fully connected layers of the binary transformer block (BTFB) are the binary normalized layers (BNFCL) presented in Algorithm \ref{alg:01}. The embedding layer is the binary embedding layer (BEMBL) presented in Algorithm \ref{alg:o4} and the normalization layer performs a simple normalization, i.e., it returns data with zero mean and unit standard deviation.

Algorithm \ref{alg:05} presents the forward propagation process in a binary transformer block (BTFB). The required inputs of Algorithm \ref{alg:05}, are the input token sequence ($seq$), the embeddings dimension ($emb\_dim$), the number of heads in the attention layer ($num\_heads$), and the number of units of the first BNFC layer ($ff\_dim$).

\begin{algorithm*}
\caption{Forward propagation calculation process in a binary transformer block (BTFB)}
\begin{algorithmic}[1]
\Require Input $seq$, embedding dimension $emb\_dim$, number of heads in the attention layer $num\_heads$, number of units of the first FCL layer $ff\_dim$
\Ensure Transformer output $output$
\State Apply binary normalized attention mechanism: 
\State \ \ \ \ $attention\_output = \texttt{BATL(emb\_dim}=emb\_dim, \texttt{num\_heads}=num\_heads)(seq, seq, seq)$
\State Add and normalize: $add\_norm = \texttt{Normalize}(seq + attention\_output)$
\State Apply binary normalized fully connected layers (BNFCL):
\State \ \ \ \ $ffn\_output = \texttt{BNFCL(units}=ff\_dim, \texttt{activation}=\text{'gelu'})(add\_norm)$
\State \ \ \ \ $ffn\_output = \texttt{BNFCL(units}=emb\_dim)(ffn\_output)$
\State Add and normalize again: $output = \texttt{Normalize}(add\_norm + ffn\_output)$
\State \Return $output$
\end{algorithmic}
\label{alg:05}
\end{algorithm*}

In Algorithm \ref{alg:05}, BNFLC(\texttt{units, activation}) represents the binary normalized fully connected layer, whose calculation process is shown in Algorithm \ref{alg:01}; note that the \textit{gelu} activation function is used in the first BNFC layer; \texttt{Normalize()} is the function that normalizes the features of each example so that it has zero mean and unit standard deviation; and \texttt{BATL(emb\_dim, num\_heads}) represents the binary multi-head attention layer. The calculation process of this attention layer is presented in Algorithm \ref{alg:06}.

Algorithm \ref{alg:06} presents the forward propagation process in a binary multi-head attention layer (BATL). The required inputs are the input token sequences $query, key$ and $value$, the causal mask ($mask$), the embeddings dimension ($emb\_dim$), and the number of heads in the attention layer ($num\_heads$). 

The functions used in Algorithm \ref{alg:06} are: \texttt{LengthOfSequence()} is a function that retrieves the length of a sequence; \texttt{Resahpe()} is a function that reallocates the elements of a tensor according to the provided shape list; \texttt{Permute()} denotes a function that swaps the axes of a tensor based on the provided order list; \texttt{Matmul()} is a function that performs tensor multiplication according to the linear algebra rules; \texttt{Where()} is the standard where function that operates conditions along all elements of a tensors; and \texttt{Softmax()} is the standard softmax function. All the other functions and terms used in Algorithm \ref{alg:06} have been defined previously.

\begin{algorithm*}
\caption{Forward propagation calculation process in a binary multi-head attention layer (BATL)}
\begin{algorithmic}[1]
\Require Input $query$, $key$ and $value$, causal mask $mask$, embedding dimension $emb\_dim$, number of heads $num\_heads$
\Ensure Final linear projection $projection$
\State Calculate number of keys: $num\_key = emb\_dim // num\_heads$
\State Apply linear projections to get Q, K, V
\State \ \ \ \ $Q = \texttt{BNFLC(units}=emb\_dim)(query)$
\State \ \ \ \ $K = \texttt{BNFLC(units}=emb\_dim)(key)$
\State \ \ \ \ $V = \texttt{BNFLC(units}=emb\_dim)(value)$
\State Get sequence length from query: $seq\_len = \texttt{LengthOfSequence}(query)$
\State Split each tensor into num\_heads to support multi-head attention:
\State \ \ \ \ $Q = \texttt{Reshape}(Q, \texttt{shape}=[-1, seq\_len, num\_heads, num\_key])$
\State \ \ \ \ $K = \texttt{Reshape}(K, \texttt{shape}=[-1, seq\_len, num\_heads, num\_key])$
\State \ \ \ \ $V = \texttt{Reshape}(V, \texttt{shape}=[-1, seq\_len, num\_heads, n\_key])$
\State Permute axis of Q, K, V to support multi-head attention
\State \ \ \ \ $Q = \texttt{Permute}(Q, \texttt{order}=[0, 2, 1, 3])$
\State \ \ \ \ $K = \texttt{Permute}(K, \texttt{order}=[0, 2, 1, 3])$
\State \ \ \ \ $V = \texttt{Permute}(V, \texttt{order}=[0, 2, 1, 3])$
\State Compute scaled dot-product attention scores: 
\State \ \ \ \ $attention\_scores = \texttt{Matmul}(Q, \texttt{Permute}(K, \texttt{order}=[0,1,3,2]))/\texttt{Sqrt}(num\_key)$
\State Apply causal mask: $scale\_dot = \texttt{Where}(mask==0, -1.0\text{e}-10, scale\_dot)$
\State Apply softmax to get attention probabilities: $attn\_prob = \texttt{Softmax}(scale\_dot, \texttt{axis}=-1)$
\State Calculate attention: $A = \texttt{Matmul}(attn\_prob, V)$
\State Reshape attention back to the original dimension: 
\State \ \ \ \ $A = \texttt{Permute}(A, \texttt{order}=[0, 2, 1, 3])$
\State \ \ \ \ $A = \texttt{Reshape}(A, \texttt{shape}=[-1, seq\_len, num\_heads*num\_key])$
\State Apply final linear projection: $projection = \texttt{BNFLC}(\texttt{units}=emb\_dim)(A)$
\State \Return $projection$
\end{algorithmic}
\label{alg:06}
\end{algorithm*}

\section{Image classification problem}
\label{sec:ConvolutionalModel}

For the image multiclass classification problem, a binary convolutional model is configured and the Food-101 dataset \cite{bossard2014food101} is used. The Food-101 dataset has a total of 101,000 images in varying resolutions with 101 categories of foods. This dataset is used for identification of types of food in a dish. The data is divided in two sets: the training data with 75,750 images and the validation data with 25,250 images.

\subsection{Configuration of the models with convolutional layers}
Algorithm \ref{alg:07}  outlines the binary convolutional model (BCVNN) for the image classification task. The inputs for the model are an image (\texttt{image}) and the filter dimension used in the convolutional layers f. The output of the model are the 101 class probabilities calculated for the input image (\texttt{probs}). In Algorithm \ref{alg:07}  the following functions are used: \texttt{BNCVL()} represents the binary normalized convolutional layer presented in \ref{alg:03}; \texttt{MAXPOOL2D()} represents a max-pooling layer; \texttt{BNFCL()} represents the binary normalized fully connected layer presented in Algorithm \ref{alg:01}; and \texttt{GLOBALAVG()} is a standard global average pooling layer that averages a three-axis tensor across the first two dimensions resulting a tensor with only one axis. All convolutional layers use \textit{relu} activation function, a stride of 1, and padding to maintain the width and height of the tensors. All max-pooling layers use $2 \times 2$ windows and stride equal to 2. The first and second binary normalized fully connected layer use \textit{relu} activation, and the output layer uses softmax activation. 

\begin{algorithm*}
\caption{Binary convolutional model used for the image classification problem (BCVNN)}
\begin{algorithmic}[1]
\Require Input $image$, filter dimension $f$
\Ensure Classes probabilities $prob$
\State First block of convolutional layers
\State \ \ \ \ $a1 = \texttt{BNCVL}(\texttt{units}=32, (f, f), \texttt{activation}='\text{relu}', \texttt{padding}='\text{same}')(image)$
\State \ \ \ \ $a1 = \texttt{BNCVL}(\texttt{units}=32, (f, f), \texttt{activation}='\text{relu}', \texttt{padding}='\text{same}')(a1)$
\State \ \ \ \ $a1 = \texttt{MAXPOOL2D}(\texttt{window}=(2,2), \texttt{stride}=(2,2))(a1)$
\State Second block of convolutional layers
\State \ \ \ \ $a2 = \texttt{BNCVL}(\texttt{units}=64, (f, f), \texttt{activation}='\text{relu}', \texttt{padding}='\text{same}')(a1)$
\State \ \ \ \ $a2 = \texttt{BNCVL}(\texttt{units}=64, (f, f), \texttt{activation}='\text{relu}', \texttt{padding}='\text{same}')(a2)$
\State \ \ \ \ $a2 = \texttt{MAXPOOL2D}(\texttt{window}=(2,2), \texttt{stride}=(2,2))(a2)$
\State Third block of convolutional layers
\State \ \ \ \ $a3 = \texttt{BNCVL}(\texttt{units}=64, (f, f), \texttt{activation}='\text{relu}', \texttt{padding}='\text{same}')(a2)$
\State \ \ \ \ $a3 = \texttt{BNCVL}(\texttt{units}=64, (f, f), \texttt{activation}='\text{relu}', \texttt{padding}='\text{same}')(a3)$
\State \ \ \ \ $a3 = \texttt{MAXPOOL2D}(\texttt{window}=(2,2), \texttt{stride}=(2,2))(a3)$
\State Fourth block of convolutional layers
\State \ \ \ \ $a4 = \texttt{BNCVL}(\texttt{units}=128, (f, f), \texttt{activation}='\text{relu}', \texttt{padding}='\text{same}')(a3)$
\State \ \ \ \ $a4 = \texttt{BNCVL}(\texttt{units}=128, (f, f), \texttt{activation}='\text{relu}', \texttt{padding}='\text{same}')(a4)$
\State \ \ \ \ $a4 = \texttt{MAXPOOL2D}(\texttt{window}=(2,2), \texttt{stride}=(2,2))(a4)$
\State Fifth block of convolutional layers
\State \ \ \ \ $a5 = \texttt{BNCVL}(\texttt{units}=256, (f, f), \texttt{activation}='\text{relu}', \texttt{padding}='\text{same}')(a4)$
\State \ \ \ \ $a5 = \texttt{BNCVL}(\texttt{units}=256, (f, f), \texttt{activation}='\text{relu}', \texttt{padding}='\text{same}')(a5)$
\State \ \ \ \ $a6 = \texttt{GLOBALAVG}()(a5)$
\State Classification layers
\State \ \ \ \ $a7 = \texttt{BNFCL}(\texttt{units}=256, \texttt{activation}='\text{relu}')(a6)$
\State \ \ \ \ $a8 = \texttt{BNFCL}(\texttt{units}=256, \texttt{activation}='\text{relu}')(a7)$
\State \ \ \ \ $prob = \texttt{BNFCL}(\text{units}=101, \texttt{activation}='\text{softmax}')(a8)$
\State \Return $prob$
\end{algorithmic}
\label{alg:07}
\end{algorithm*}

\subsection{Convolutional models training}
To verify whether the number of parameters of the models influences training stability and performance of the binary models, two models with different filter dimensions are configured and trained: $3 \times 3$ and $5 \times5$. The model with $3 \times 3$ filters has 5,132,165 parameters and the model with $5 \times5$ filters has 13,505,925 parameters. 

To verify if the binary models are effective, two models with float 32-bit parameters ("standard" models) with the same configurations of the binary models are also configured and trained. In these ``standard" models dropout layers are introduced after the first and second fully connected layers with dropout rates of 0.4 and 0.3 respectively. Dropout is necessary in the standard models to prevent excess overfitting.

In all models, connection weights and biases are initialized using the standard methods: Glorot Uniform for weights and zeros for biases. No regularization methods or parameter constraints are applied in the binary models and only dropout are used in the standard models. Table \ref{tab:Hyp_ConvModel} presents the hyperparameters used for training the convolutional models.

\begin{table}[h]
\centering
\caption{Hyperparameters used for training the convolutional models.}
\begin{tabular}{|l|l|}
\hline
Hyperparameter & Value \\
\hline
Image resolution & $256 \times 256 \times 3$ \\
Cost function & Categorical cross entropy \\
Metrics & Accuracy \\
Optimization method & Adam \\
Batch size & 64 \\
Learning rate schedule & Warmup and decay \\
Maximum learning rate & 0.0001 \\
Warmup steps & 20 \\
Decay steps & 1100 \\
Number of epochs & 1000 \\
\hline
\end{tabular}
\label{tab:Hyp_ConvModel}
\end{table}

It is observed that a considerable number of training epochs are needed for the cost function to completely converge during training.

\subsection{Results obtained with the models with convolutional layers}
\begin{figure*}
\centering
\includegraphics[width=0.9\textwidth]{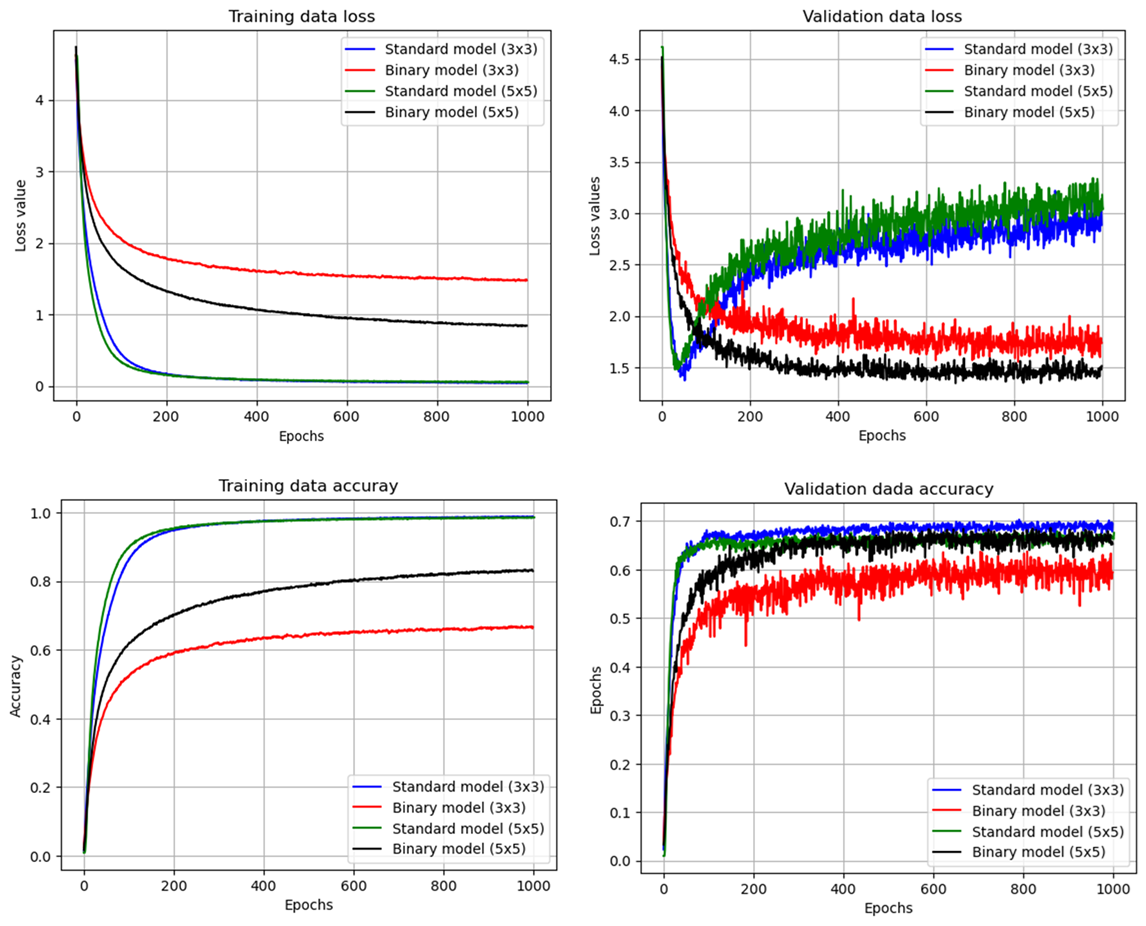}
\caption{Training results of image classification problem with the convolutional models.}
\label{fig:ConvolutionalModels}
\end{figure*}

In Figure \ref{fig:ConvolutionalModels}, the training results are displayed for the convolutional models. The results from the standard models with 32-bit parameters are included as a benchmark for the desired performance. It is important to note that multiple training tests were conducted for all models, and all results are very similar. These results are summarized in Table \ref{tab:Results_ConvolutionalModels} that presents the best results obtained during training for each model.

\begin{table*}[h]
\centering
\caption{Summary of the results of the image classification problem with the convolutional models.}
\begin{tabular}{|l|l|l|l|l|}
\hline
Model & Training loss & Validation loss & Training accuracy & Validation accuracy \\
\hline
Standard model with $3 \times3$ filter & 0.0369 & 1.37 & 0.989 & 0.703 \\
Binary model with $3 \times3$ filter & 1.46 & 1.55 & 0.670 & 0.637 \\
Standard model with $5 \times5$ filter & 0.0495 & 1.48 & 0.986 & 0.679 \\
Binary model with $5 \times5$ filter & 0.836 & 1.35 & 0.834 & 0.686 \\
\hline
\end{tabular}
\label{tab:Results_ConvolutionalModels}
\end{table*}

Analyzing the training results of the convolutional models shown in Figures 1 and Table 2, the following observations can be made:
\begin{itemize}
\item The binary models are capable to train without any kind of instability and their performance is almost equal to the standard models;
\item The standard models learn more rapidly than the binary models, i.e., they need fewer epochs for training;
\item The standard models present strong overfitting while the binary models do show overfitting;
\item The accuracies for the validation data of the standard models are slightly better than the ones of the binary models;
\item The binary model with 5x5 filters presents better performance than the 3x3 filters binary model;
\item The results of the binary models are very good considering that they have only 1-bit parameters.
\end{itemize}

It is important to observe that the binary normalization layers are effective to solve the problems of training instabilities and low accuracy of models with binary parameters. According to the study performed by \cite{Cabral2025}, that analyzed the impact of low-resolution parameters on the performance of neural networks, models with binary parameters are not able to train effectively.

\section{Language decoder problem}
\label{sec:LanguageModel}

For the language decoder problem, a binary transformer model is configured and the WikiText-103-raw dataset is used. This dataset was created by Salesforce Research \cite{merity2016}. The \texttt{WikiText-103-raw} dataset is primarily sourced from English Wikipedia. Specifically, it was created from high-quality Wikipedia articles to provide clean and representative data for training language models. It includes 25,000 carefully selected Wikipedia articles, containing around 103 million words. The "raw" version preserves the original punctuation and basic formatting, unlike the tokenized version. The dataset was pre-processed in the form of sentences performing 782,208 examples. The data was divided into training dataset with 95\% of the examples and validation dataset with the rest 5\%.

The text is tokenized using the \texttt{WordPieceTokenizer} \cite{song2021wordpiece} which uses a sub-word strategy. Its vocabulary size is 30,522, and any token not appearing in the vocabulary is replaced by [UNK] ("unknown").

\subsection{Configuration and training the language decoder}
Algorithm \ref{alg:08}  outlines the binary language decoder model (BLM). The inputs for the model are the sequence of tokens ($seq$), the maximum sequence length $max\_len$, the embedding dimension $emb\_dim$, the number of attention heads $num\_heads$, the vocabulary size $vocab\_size$, the numbers of units in the MLP head $mlp\_units\_0$ and $mlp\_units\_1$. The output of the model are the $vocab\_size$ probabilities calculated for the next token ($probs$). In Algorithm \ref{alg:08}  the following functions are used: \texttt{BEMB()} represents the binary embedding layer presented in Algorithm \ref{alg:03}; \texttt{Normalize()} is the function that normalizes the features of each example so that it has zero mean and unit standard deviation; \texttt{BTFB()} is the transformer block presented in Algorithm \ref{alg:05}; \texttt{BNFCL()} represents the binary normalized fully connected layer presented in Algorithm \ref{alg:01}. The activation functions of the MLP head layers are \textit{gelu} and for the last layer is \textit{softmax}.

\begin{algorithm*}
\caption{Binary language decoder model (BLM)}
\begin{algorithmic}[1]
\Require Input token sequences $seq$, maximum sentence length $max\_len$, embedding dimension $emb\_dim$, number of attention heads $num\_heads$, vocabulary size $vocab\_size$, number of units in the mlp head layers $mlp\_units\_0$ and $mlp\_units\_1$
\Ensure Classes probabilities $probs$
\State Embedded coding of the token sequences: $embs = \texttt{BEMB}(max\_len, emb\_dim, vocab\_size)(seq)$
\State Embedding normalizations: $x = \texttt{Normalize}(embs)$
\State Pass through a sequence of transformer blocks:
\For{$i$ from 1 to $num\_blocks$}
    \State $x = \texttt{BTFB}(emb\_dim, num\_heads, \texttt{ff\_dim}=2*emb\_dim)(x)$
\EndFor
\State Process transformer output with fully connected layers (MLP head)
\State \ \ \ \ $features = \texttt{BNFCL}(\texttt{units}=mlp\_units\_0, \texttt{activation}='\text{gelu}')(x)$
\State \ \ \ \ $features = \texttt{BNFCL}(\texttt{units}=mlp\_units\_1, \texttt{activation}='\text{gelu}')(features)$
\State Final fully connected layer to calculate the probabilities
\State \ \ \ \ $probs = \texttt{BNFCL}(\texttt{units}=vocab\_size, \texttt{activation}='\text{softmax}')(features)$
\State \Return $probs$
\end{algorithmic}
\label{alg:08}
\end{algorithm*}

To verify the influence of the number of parameters on training stability and performance, two binary transformer neural networks are configured with different number of parameters. Table \ref{tab:Hyp_LM} presents the hyperparameters used to configure the binary language models. The small model has about 154.4 million binary parameters and the large model has about 332.8 million.

\begin{table}[h]
\centering
\caption{Hyperparameters used to configure the language models.}
\begin{tabular}{|l|l|l|}
\hline
Hyperparameter & Small model & Large model \\
\hline
Maximum sentence length & 256 & 256 \\
Number of transformer blocks & 12 & 16 \\
Embedding dimension & 768 & 1024 \\
Number of attention heads & 16 & 16 \\
Number of units in the MLP head & 4096 - 2048 & 8192 - 4096 \\
Vocabulary size & 30522 & 30522 \\
Total number of parameters & 154.4 million & 332.8 million \\
\hline
\end{tabular}
\label{tab:Hyp_LM}
\end{table}

To verify if the binary models are effective, a model with float 32-bit parameters ("standard" model) with the same configurations of the small binary model is also configured and trained. In the "standard" model the normalize layers has trainable parameters to adjust the best mean and standard deviation of the data, therefore, the standard model has a slightly larger total number of parameters than the equivalent binary model.

In all models, connection weights and biases are initialized using the standard methods: Glorot Uniform for weights and zeros for biases. No dropout is used in any model and no regularization methods or parameter constraints are applied in all models. Table \ref{tab:HYP_LDM} presents the hyperparameters used for training the language models.

\begin{table}[h]
\centering
\caption{Hyperparameters used for training the language decoder models.}
\begin{tabular}{|l|l|}
\hline
Hyperparameter & Value \\
\hline
Cost function & Categorical cross entropy \\
Metrics & Accuracy and Perplexity \\
Optimization method & AdamW \\
Learning rate & $1.0 \times 10^{-5}$ \\
Number of epochs & 100 \\
\hline
\end{tabular}
\label{tab:HYP_LDM}
\end{table}

\subsection{Results obtained with the language decoders}
\begin{figure*}
\centering
\includegraphics[width=0.9\textwidth]{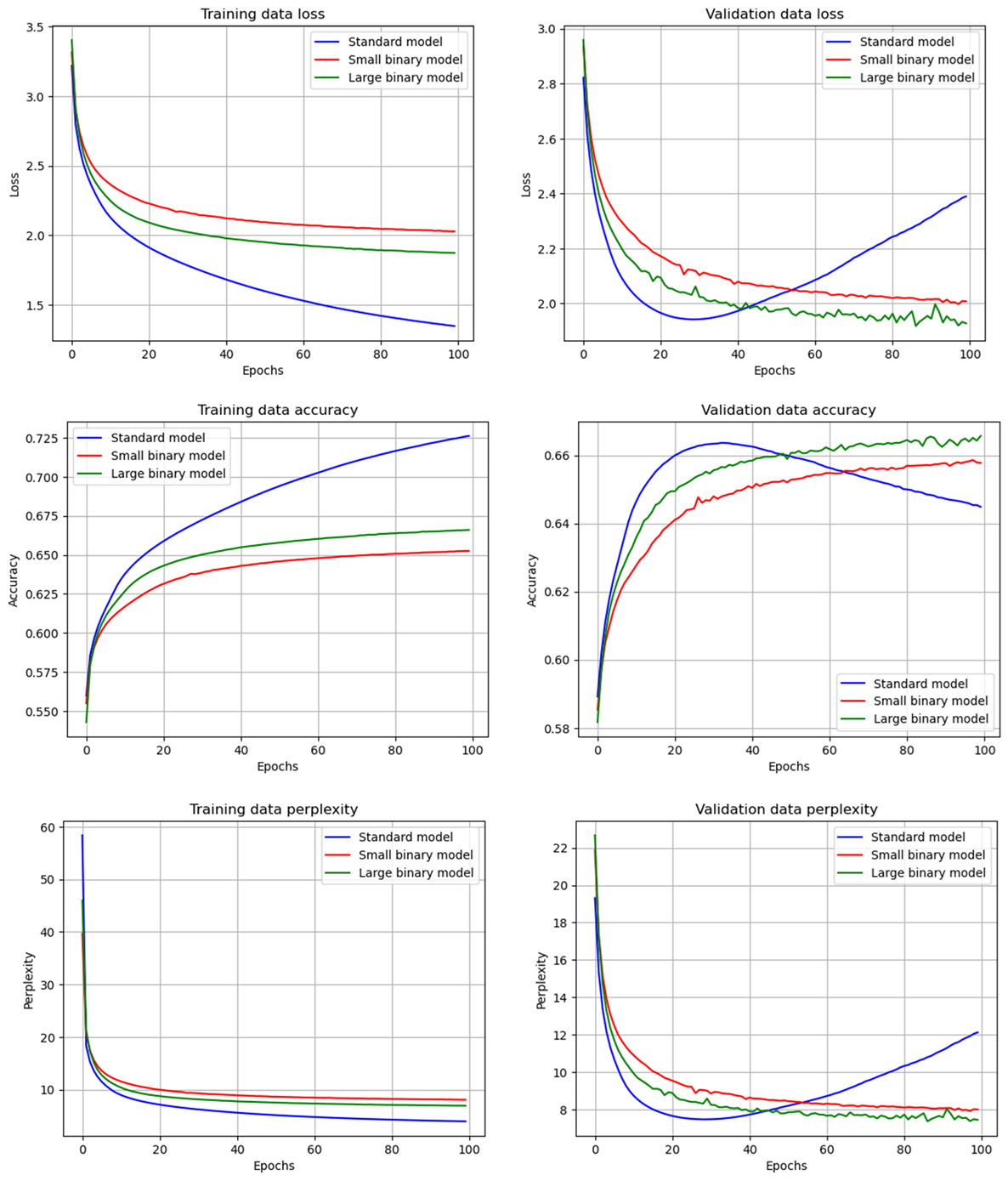}
\caption{Training results of the language decoders.}
\label{fig:LanguageDecoders}
\end{figure*}

In Figure \ref{fig:LanguageDecoders} the training results are displayed for the language decoder models and in Table \ref{tab:Results_LDM} these results are summarized. The values given in Table \ref{tab:Results_LDM} are the best results obtained during training. The results from the standard language decoder with 32-bit parameters are included as a benchmark for the desired performance. It is important to note that multiple training tests were conducted for all models, and all results are very similar.

\begin{table*}[h]
\centering
\caption{Summary of the language decoder models results}
\label{tab:Results_LDM}
\begin{tabularx}{\textwidth}{@{}lXXXXXX@{}}
\toprule
Model & Training loss & Validation loss & \multicolumn{1}{c}{Training accuracy} & \multicolumn{1}{c}{Validation accuracy} & Training perplexity & Validation perplexity \\
\midrule
Standard model & 1.35 & 1.94 & 0.726 & 0.664 & 3.98 & 7.47 \\
Small binary model & 2.03 & 1.99 & 0.653 & 0.659 & 8.08 & 7.92 \\
Large binary model & 1.87 & 1.91 & 0.666 & 0.666 & 6.95 & 7.47 \\
\bottomrule
\end{tabularx}
\end{table*}

Analyzing the training results of the decoders models shown in Figure 2 and Table 5, the following observations can be made:
\begin{itemize}
\item The binary models are capable to train without any kind of instability and their performance are similar to the standard model;
\item The standard model shows a large overfitting while none of the binary models present overfitting;
\item The results for the validation data of the binary models are very similar of the results of the standard model;
\item The results of the large binary model are slightly better than the results of the small binary model and are equivalent to the best results of the standard model. Note that the large binary model has roughly the double number of parameters of the small binary model and the standard model;
\item The results of the binary models are very good considering that they have only 1-bit parameters.
\end{itemize}

Again, it is important to observe that the binary normalization layers are effective to solve the problems of training instabilities and low accuracy of models with binary parameters.


\section{Conclusions}
\label{sec:Conclusions}

In this work, a novel type of neural network model is developed, in which all parameters have only a single bit. In this new class of models, all layer parameters — including kernel weights and biases — are restricted to 1-bit resolution. These models are built using a type of layer referred to as a binary normalized layer. Binary normalized layers can be implemented in various architectures, such as fully connected, convolutional, or attention-based layers.

To show the effectiveness of the binary normalized layers, two different types of problems are solved: a multiclass image classification problem, and a language decoder to predict the next token of a sequence. A convolutional binary model is configured to solve the image classification problem and a binary transformer model is configured to solve the language problem. Standard 32-bit float parameters models with the same configuration of the binary models are used to provide benchmarks for the results.

The model used for image classification includes convolutional and fully connected layers. Two versions of this model are configured and tested, differing only in the number of parameters. Similarly, the model used for the language decoding task consists of transformer blocks with multi-head attention layers, and again, two versions are tested, differing in the number of transformer blocks and total parameters.

Results show that models using binary normalized layers achieve almost the same performance to that of the corresponding models with 32-bit floating-point parameters. Moreover, the binary models do not exhibit instability during training—an important result, since training instability is a major concern in low-resolution models. As observed by \cite{Cabral2025}, traditional models with binary parameters typically fail to train effectively due to such instability.

As expected, increasing the number of parameters in the binary models improves their performance without causing overfitting. This is a key advantage of the proposed model type, allowing it to reach performance levels similar to those of 32-bit models.

It is important to emphasize that no effort was made to fine-tune complex architectures for maximal performance or to eliminate overfitting entirely. The primary goal of this study is to introduce the binary normalized model and demonstrate its ability to generalize and perform comparably to standard 32-bit models. For this purpose, a direct comparison with equivalent 32-bit models is sufficient.

Binary normalized layers make it possible to build models that use 32 times less memory than conventional networks while delivering equivalent performance. Furthermore, these layers can be efficiently implemented on standard hardware using 1-bit arrays and do not require dedicated electronic components. This new type of layer opens the door to large-scale neural networks with drastically reduced memory requirements, making them deployable on simple and inexpensive hardware such as mobile devices or CPUs alone.

In addition, 1-bit parameter models offer the significant advantage of enabling the development of more complex architectures with more processing units while still using far less memory than traditional 32-bit models or even those using 8-bit quantization. These low-resolution networks have the potential to support the deployment of large language models on embedded devices.

Future work will focus on the implementation of binary normalization layers using single-bit arrays operations, as well as on quantizing layer activations to 8 or 16-bit precision. These improvements are expected to further enhance the efficiency and performance of the binary neural network models.

\bibliographystyle{unsrtnat}
\bibliography{references}  

\end{document}